\documentclass[pmlr]{jmlr}
\pdfoutput=1

\usepackage{longtable}

\usepackage{booktabs}
\usepackage[load-configurations=version-1]{siunitx} 

\usepackage{adjustbox}
\usepackage{amsmath}
\usepackage{bbm}
\usepackage{amsfonts}
\usepackage{wrapfig}
\usepackage{amssymb}
\usepackage{multirow}
\usepackage[title]{appendix}
\usepackage{xcolor}
\usepackage{caption}  
\usepackage{subcaption} 

\makeatletter
\def\set@curr@file#1{\def\@curr@file{#1}} 
\makeatother

\theorembodyfont{\upshape}
\theoremheaderfont{\scshape}
\theorempostheader{:}
\theoremsep{\newline}

\jmlrvolume{149}
\jmlryear{2021}
\jmlrworkshop{Machine Learning for Healthcare}

\title{An Interpretable Framework for Drug-Target Interaction with Gated Cross Attention}

\author{\Name{Yeachan Kim}
      \Email{yeachan@deargen.me}\\ 
      \Name{Bonggun Shin}
      \Email{bonggun.shin@deargen.me}\\ 
      \addr Deargen Inc., Seoul, South Korea}

\begin{document}

\maketitle

\begin{abstract}
\textit{In silico} prediction of drug-target interactions (DTI) is significant for drug discovery because it can largely reduce timelines and costs in the drug development process. Specifically, deep learning-based DTI approaches have been shown promising results in terms of accuracy and low cost for the prediction. However, they pay little attention to the interpretability of their prediction results and feature-level interactions between a drug and a target. In this study, we propose a novel interpretable framework that can provide reasonable cues for the interaction sites. To this end, we elaborately design a \textit{gated cross-attention} mechanism that crossly attends drug and target features by constructing explicit interactions between these features. The gating function in the method enables neural models to focus on salient regions over entire sequences of drugs and proteins, and the byproduct from the function, which is the attention map, could serve as interpretable factors. The experimental results show the efficacy of the proposed method in two DTI datasets. Additionally, we show that gated cross-attention can sensitively react to the mutation, and this result could provide insights into the identification of novel drugs targeting mutant proteins.

\end{abstract}

\section{Introduction}
Predicting the interactions between drugs and targets is a critical step in drug discovery. The importance is more evident in repurposing (or repositioning) already-approved drugs for fast-spreading and epidemic diseases (e.g., COVID-19) because the approval process of a new drug requires significant development timelines and overall costs. Recently, deep neural networks have shown promising results on this problem \citep{ozturk2018deepdta,shin2019self}, demonstrating that they can work well in real-world scenarios \citep{beck2020predicting,choi2020target}.

Most recent studies for predicting the interaction focus on representation learning for drugs and targets. For example, \cite{shin2019self} used a transformer \citep{vaswani2017attention} to effectively capture salient features from molecules and further improved the generalization capability through a pre-training phase on a large-scale database (e.g., PubChem \citep{kim2019pubchem}). \cite{nguyen2019graphdta} replaced the sequential representations of drugs with a molecular graph using graph convolutional networks (GCNs). Although advancements in representation learning have improved the performance, most studies pay little attention to $i)$ the interpretability of their prediction results and $ii)$ the interaction of drug and target features.

To further improve the current approaches, we propose a \textit{gated cross-attention} (GCA), which is a novel interpretable and interaction framework that can provide the interpretability of the prediction results through an explicit interaction between the drug and target features. Interpretability is served through the lens of attention to drugs and targets. In other words, the higher attention score at the specific parts of the drug and target, the more significant they are for prediction. To that end, we redesign the previous attention mechanism as a \textit{gating function} so that the significant parts for the prediction easily affect to the prediction. Furthermore, unlike the previous attention mechanism that typically assigns all credits to all inputs, GCA sparsely attends a small portion of inputs to encourage the neural models to focus on salient parts over sequences.

To validate the efficacy of GCA, we apply the proposed method to existing neural models (e.g., convolutional neural networks (CNNs), GCNs) because the role of our method is to put current methods on a more solid footing by obtaining interpretability on their prediction. The comprehensive results clearly demonstrate that the proposed interaction networks not only provide the interpretable cues for the prediction by showing significant parts in drugs and targets, but also achieve performance improvement owing to the explicit interaction between features. We also conduct two case studies on mutations and possible interactions between drugs and targets and it can be the pharmaceutical use cases of the proposed method.

\textbf{Clinical Relevance}: The proposed method can help design new drugs by proposing a binding affinity with important regions of drugs and targets. It can also be used in drug design targeting specific mutant proteins by checking whether the attention in the mutation sites is activated differently.

\subsection*{Generalizable Insights about Machine Learning in the Context of Healthcare}

Due to the large length of protein sequences, effectively handling protein representation is difficult. In this case, focusing salient parts over the large sequence is significant, and the sparse attention mechanism can alleviate the problem. There are several tasks taking two inputs as an input, such as drug-drug interaction and protein-protein interaction. Cross-attention mechanism over feature extraction layers could improve the performance in tasks while providing interaction interpretability.





\section{Related Work}

\subsection{Drug Target Interaction} 
In silico prediction of DTI has been getting attention since it could provide useful guidance to drug discovery with reduced overall costs. The major approaches to DTI are three-folds which are molecular docking \citep{trott2010autodock,luo2016molecular}, similarity-based \citep{pahikkala2015toward,he2017simboost} and deep learning-based approaches. In this paper, we mainly focus on deep learning-based methods since they could provide the results promptly and achieve reasonably accurate results compared to other approaches \citep{ozturk2018deepdta}. In the beginning, several works utilized deep neural networks over character representations of drugs and targets. For drugs, SMILES (Simplified Molecular Input Line Entry System) format is utilized to represent chemical compound, and protein sequence (i.e., amino acids) is represented by FASTA \citep{lipman1985rapid}. Based on such character representations, \cite{ozturk2018deepdta} proposed DeepDTA that utilizes convolutional neural networks (CNNs) to extract locally salient features from drug and target, and it achieved better performance than similarity-based approaches. \cite{shin2019self} used the same character representation with DeepDTA, but utilized the self-attention-based transformer \citep{vaswani2017attention} to consider a long-range relationship between atoms. Furthermore, they performed transfer learning to further generalize the molecule representations by pre-training the transformer on a large-scale database. Instead of character representations, \citep{ozturk2019widedta} represented drug and target as words that correspond to substructure in compounds. Beyond such string representations, utilizing graph structures of drugs shows promising performance than others \citep{lim2019predicting,nguyen2019graphdta,zhao2020identifying}. 

The most related work to ours is \cite{gao2018interpretable}. It shares a similar strategy with ours in the aspect of attending to two inputs. However, our method adopts a multi-head attention mechanism in the transformer. It allows neural models to attend to different aspects of information that lie in input features. Thus, it serves more options to interpret the prediction results. For example, we found out that one of the heads from GCA is more committed to predicting the binding site than other heads (detailed in Section 5.2). Beyond the differences in architecture, the proposed model constraints the attention map to be sparse, which guides the model to more focus salient parts in inputs.


\subsection{Attention Method in Neural Networks}
Attention mechanism was invented to improve the quality of machine translation by aligning two different representations (e.g., source and target languages) \citep{bahdanau2014neural,luong2015effective}. There are two advantages in utilizing attention: First, it allows neural models to effectively consider a long-range relationship between features and it leads to the improved task performance \citep{yang2016hierarchical,anderson2018bottom}. Another advantage of the attention mechanism is that it could provide the interpretability of the model prediction \citep{seo2017interpretable,annasamy2019towards}. In the interaction task of different domains, several works utilize the attention mechanism to provide interpretability. For example, \cite{santos2016attentive} utilized pooling operation over the interactive attention maps between question and answer for an answer selection task. Similarly, \cite{gong2017natural} used densely-connected networks over attention maps to extract locally salient interactions between premise and hypothesis for a natural language inference task. For a molecule optimization task, \cite{shin2021controlled} utilized a self-attention-based transformer to optimize multiple properties of a given molecule.

For drug-target interaction, most works have shown the strengths of the attention mechanism at generating better representations. However, they pay little attention to the interpretability and interaction between two inputs. In this work, we focus on the above properties to improve the task performance and provide reasonable cues for the factors to the interaction binding, which may have broad pharmaceutical applications. Note that our method is parallel research lines with other works. Thus, we believe that existing methods could benefit from ours.

\section{Gated Cross-Attention Networks} 

\begin{figure*}[t]
    \centering
    \includegraphics[width=0.9\textwidth]{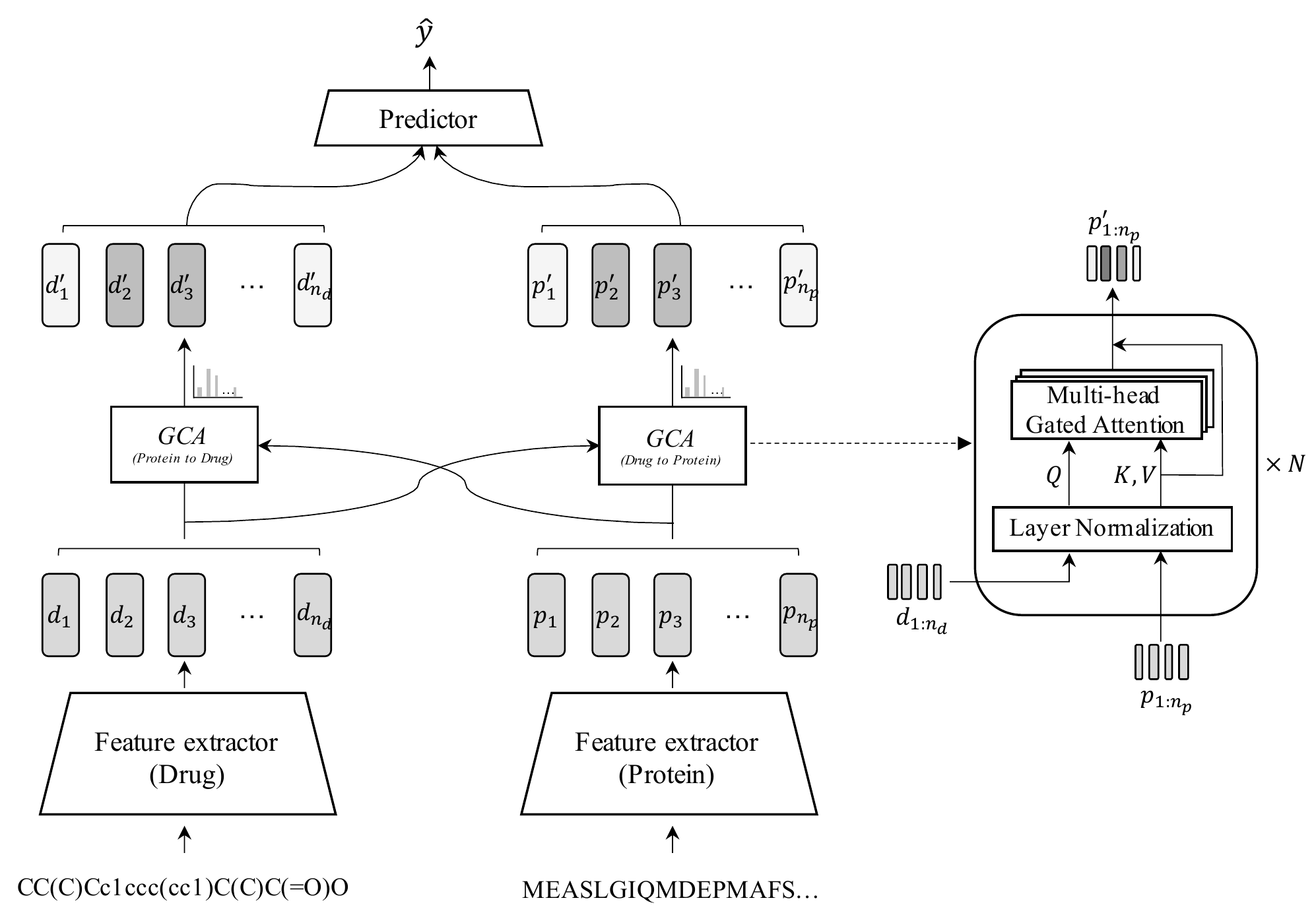}
    \caption{Overview of gated cross attention (GCA) networks and the detailed procedures when deriving the protein attention (right). GCA explicitly constructs the interaction between drug and target features, which are obtained from each feature extractor, using multi-head gated attention. The attention map, which is the by-product of the explicit interaction, serves as an interpretable factor for the prediction results.}
    \label{model}
\end{figure*}

In this section, we elaborate the proposed method which is denoted as gated cross attention (GCA). Before delving into the method, we first describe the background of drug-target interaction and formally define the task (Section \ref{dti}). Then, we introduce the attention mechanism in the transformer with some negative observation at the interaction task (Section \ref{mha}). Lastly, we detail gated cross attention (Section \ref{mhga}). The overall procedures are straightforward and illustrated in Figure \ref{model}. The main strategy of GCA is to generate drug and target features by considering counterpart information. The name \textit{\textbf{cross}} in our method comes from the directionality of  the attention in our method

\subsection{Drug-Target Interaction}\label{dti}
The task of this paper is to predict the affinity between a drug and a protein (also noted as target). The affinity can be measured as either scores or categories, the task of which is then regression or classification, respectively. Since predicting affinity scores is much beneficial in the process of drug discovery \citep{ozturk2018deepdta}, we focus on regression-based modeling. Formally, the inputs are two kinds: a molecule represented by the SMILES \citep{weininger1988smiles} sequence or the molecular graph, and a protein represented by the FASTA \citep{lipman1985rapid} sequence. A FASTA sequence is a series of various amino acids (also noted as residues).

The typical neural frameworks for drug-target interaction consist of two major components: \textbf{1)} two feature extraction layers for drugs and targets, and \textbf{2)} a single fully-connected layer to predict binding affinity scores based on the resultant features from \textbf{1)}. Let the drug and target features from each feature extractor be $d_{1:n_d}$ and $p_{1:n_p}$ where $n_d, n_p$ are the sequential length of each feature and the subscript indicates the sequential range of features, we formally define the task as follows:
\begin{equation} \label{pool}
    r_d = pooling(d_{1:n_d}) , \text{  } 
    r_p = pooling(p_{1:n_p}),
\end{equation}

\begin{equation}
    \hat{y} = \phi(r_d, r_p;\theta_i).
\end{equation}
where $pooling(\cdot)$ is the pooling methods such as global max pooling and average pooling, which depends on the architecture. $\hat{y}$ is the predicted binding affinity score by an interaction layer $\phi(\cdot)$ with its parameter $\theta_i$. Most works used simple feed-forward networks for an interaction function. In this work, we enhance the interaction parts $\phi(\cdot)$ by explicitly making the interaction between drugs $d_{1:n_d}$ and targets $p_{1:n_p}$.

\subsection{Cross Attention}
To explicitly consider the interaction, we propose a \textit{cross-attention} that considers protein representations to attend the drug representations and vice versa. We realize this by cross-attending each feature by its counterpart representations. We formalize this as follows:
\begin{equation}
    d^{'}_{1:n_d} = g_{p \rightarrow d}(d_{1:n_d}, p_{1:n_p}; \theta_{p \rightarrow d}),
\end{equation}
\begin{equation}
    p^{'}_{1:n_p} = g_{d \rightarrow p}(p_{1:n_p}, d_{1:n_d}; \theta_{d \rightarrow p}).
\end{equation}
where $g(\cdot)$ indicates an attention function that considers counterpart information. For example, $g_{p \rightarrow d}$ takes protein features to attend drug features with its learnable parameters $\theta_{p \rightarrow d}$ and returns the attended features of $d_{1:n_d}$. The attention weights, which are the byproducts of $g(\cdot)$, can serve as interpretable factors because the weights exhibit the most contributing sequences (i.e., amino acids and atoms) to predict affinity. In the following section, we will describe the functions are suitable for $g(\cdot)$.

\subsubsection{Multi-head Attention}\label{mha}
As an attention function $g$, we follow multi-head attention in the transformer. There are two different types of attention: \textbf{1)} encoder attention and \textbf{2)} decoder attention since the transformer was constructed for a sequence-to-sequence task (e.g., machine translation). We only describe the process of calculating attention for drug features $d_{1:n_d}$ because the attention for a protein is symmetric with that of drugs.

\subsection*{Encoder Attention}
The key idea of the encoder attention is to attend inputs by using the same inputs but differently transformed, which is also called \textit{self-attention}. This attention is repeatedly performed with different heads and layers. For simplicity, we show the process of a single layer with a single head. 

We first construct \textit{query}, \textit{key}, and \textit{value} by transforming the same inputs $d_{1:n_d}$ by weight matrices $W_Q$, $W_K$, and $W_V$. These features are denoted as $Q_d$, $K_d$, and $V_d$, respectively and share the same dimensionality with the inputs (i.e., $Q,K,V \in \mathbbm{R}^{n_d \times b}$). The attention weights $A \in \mathbb{R}^{n_d \times n_d}$ are computed based on the query and key:
\begin{equation}\label{MHA_1}
    A = \text{softmax}(\frac{Q_{d}K_{d}^{T}}{\sqrt{b}}),
\end{equation}
where $b$ is the dimensionality of input features and it serves as temperature to relax the distribution of the attention. Using the attention weights $A$ and the values $V_d$, we compute attended inputs as follows:
\begin{equation}\label{MHA_2}
    V_{d}^{'} = AV_{d}.
\end{equation}
Here, $A$ can be utilized as interpretable factor to the prediction. The above encoder attention is widely utilized at feature extraction layers in various domains, such as image \citep{cornia2020meshed}, natural language processing \citep{devlin2018bert}, and drug-discovery \citep{shin2019self,shin2021controlled}.

\subsection*{Decoder Attention}
Decoder attention is slightly different from the self-attention. This attention is used at the decoder in the transformer, and the goal is to align decoder features with encoder features. To align two different features, it uses the counterpart features $p_{1:n_p}$ instead of its inputs $d_{1:n_d}$ to construct the key and value, which are denoted as $K_p$ and $V_p$, and the attended features in Eq. \ref{MHA_2} are obtained as follows:
\begin{equation}\label{MHA_3}
    V_{d}^{'} = \text{softmax}(\frac{Q_{d}K_{p}^{T}}{\sqrt{b}})V_{p},
\end{equation}

\begin{figure}[!ht]
  \centering
  \subcaptionbox{Decoder Attention}{\includegraphics[width=.47\textwidth]{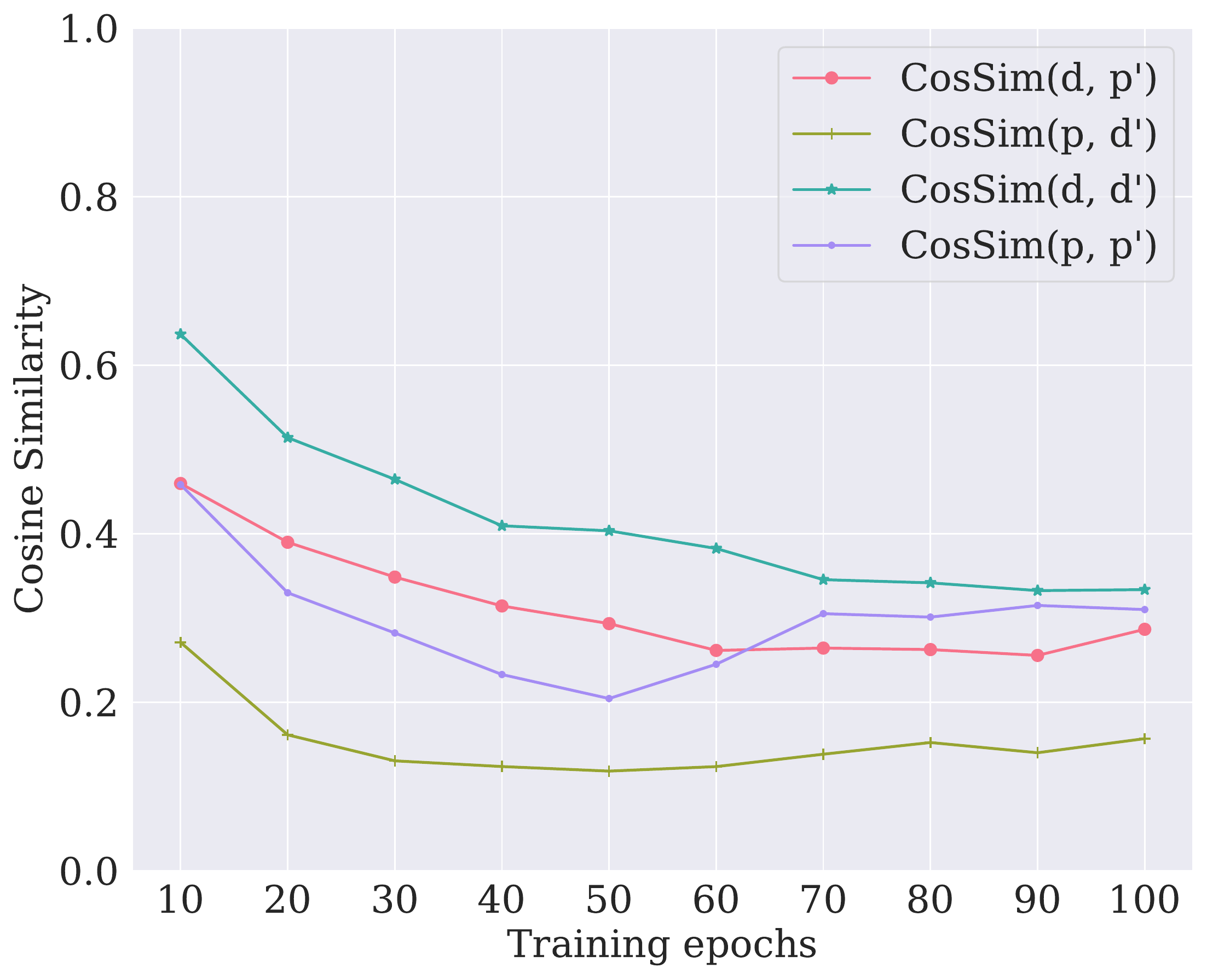}} 
  \subcaptionbox{Gated Cross Attention}{\includegraphics[width=.47\textwidth]{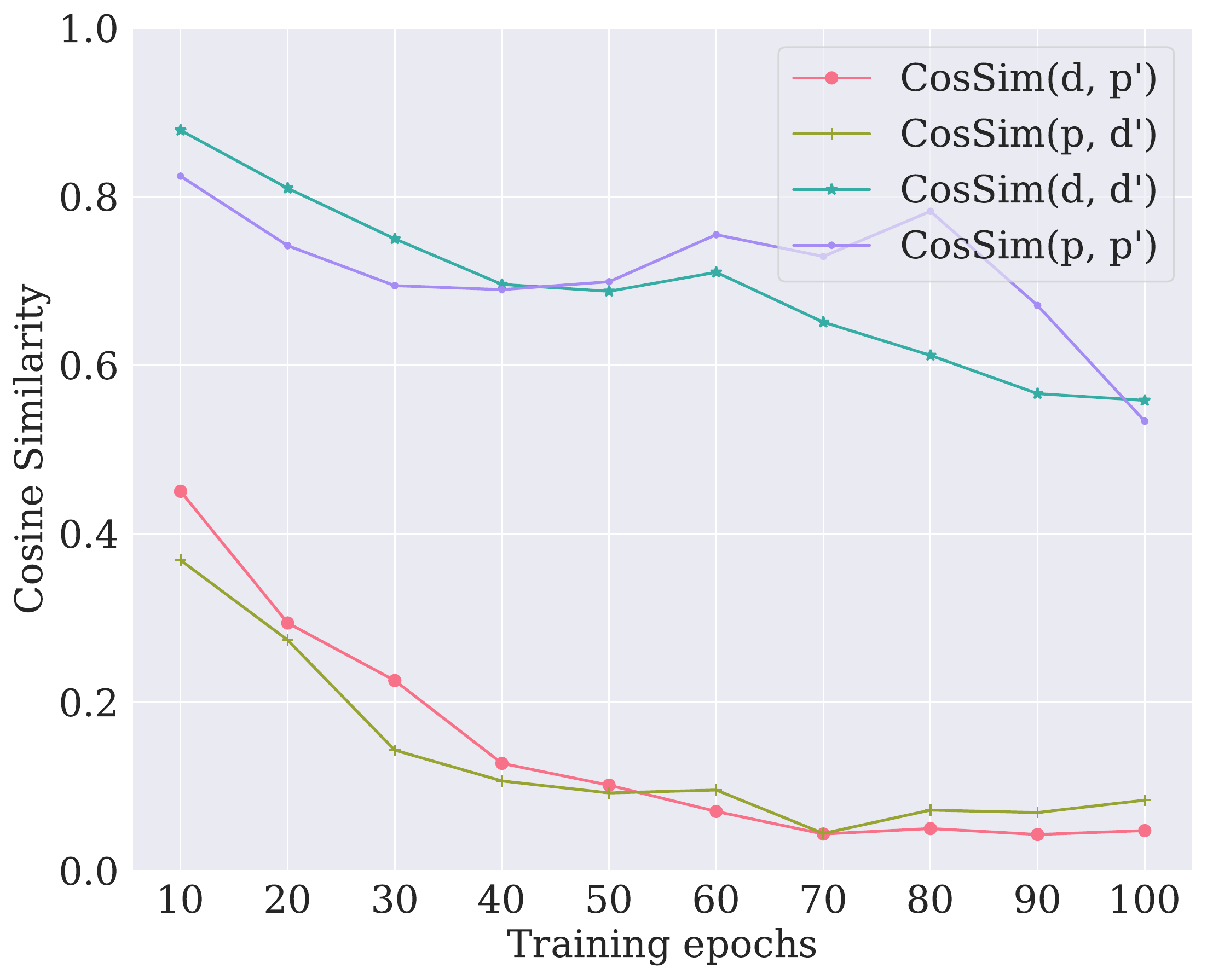}} 

\caption{Pair-wise similarity between drugs ($d, d^{'}$) and protein features ($p, p^{'}$) over different training epochs when using different attention mechanism. It shows that utilizing decoder attention does not generate discriminate features between drug and target.}
\vspace*{-0.2in}
\label{similarity}
\end{figure}

The decoder attention can be used as a $g_{p} \rightarrow {d}(\cdot)$ in the cross attention. However, we empirically found that the decoder attention results in highly mixed representations between the two features. Figure \ref{similarity}(a) shows the pairwise feature similarity between drugs ($d, d^{'}$) and proteins ($p, p^{'}$) which are inputs and outputs from the decoder attention-based $g(\cdot)$ function. We observe that the protein features $p^{'}$ from the decoder attention are more similar to drug features $d$ than its input features $p$ over several training ranges (e.g., 3k-18k and near 30k) while the attended features for drugs are more similar to its input than protein features. In other words, a specific modality could dominate the other, and we found that such cases lead to poor performance in the task. These results are derived from the method of attention weighting in Eq. \ref{MHA_3} because the attention maps $A$ are applied to counterpart features rather than its input features. This is the intended design for many sequence-to-sequence tasks, where the goal is to effectively align the target and source. However, we argue that this should be redesigned for interaction tasks.


\subsubsection{Multi-head Gated Attention}\label{mhga}
For the interaction task, we redesign the decoder attention. It shares the same strategy of multi-head attention with the decoder. However, it is vastly different in that \textbf{1)} It uses context-level attention instead of token-level attention \textbf{2)} The attention is applied in the manner of \textit{gating} mechanism in the variants of recurrent neural networks \citep{chung2014empirical,bradbury2016quasi}. The attention weights $a \in \mathbbm{R}^{n_d}$ for drug features are calculated as follows:
\begin{equation}\label{MHGA_1}
        a = \frac{1}{n_p}\sum_{i=1}^{n_p} \text{softmax} (\frac{Q_{p}K_{d}^{T}}{\sqrt{b}})_{:, i},
\end{equation}
In comparison to the decoder attention, the query and key inputs are reversed, and the attention map is aggregated by counterpart information, yielding a context-level attention vector rather than a token-wise attention matrix (Eq. \ref{MHA_1}). The attention vector $a$ serves as a gating function in calculating the attended features with the values $V_d$:
\begin{equation}\label{MHGA_2}
        V_{d}^{'} = \text{softmax}(a) \odot V_d.
\end{equation}
where $\odot$ is an element-wise multiplication operation\footnote{If the dimensions do not match, the operation is performed in a broadcasting manner. We follow NumPy \citep{harris2020array} convention (https://numpy.org/doc/stable/user/theory.broadcasting.html)}. After applying attention, we add a residual connection between attention layers, which creates good local minima \citep{yun2019deep}, and follow pre-normalization \citep{xiong2020layer}, as illustrated in Figure \ref{model} (right). 

The attended values $V^{'}$ are $d^{'}_{1:n_{d}}$, and these features are pooled over the length $n_{d}$ in Eq. \ref{pool}. For the protein side, the attended features $p^{'}_{1:n_{p}}$ are symmetrically computed using the above procedure. Lastly, the concatenated features of a drug and a target are fed to multi-layered feed-forward networks to predict the binding affinity. To train the overall networks for the drug-target interaction task, we minimize the mean squared error (MSE) between predicted affinity scores and ground-truth scores. Figure \ref{similarity}(b) shows that gated attention generates highly discriminative features while considering counterpart information.


\subsubsection{Sparse Attention}
Our attention mechanism assigns probabilities (i.e., $a$ in Eq \ref{MHGA_1} and \ref{MHGA_2}) over all elements. However, most of them have a small but non-zero probability, which could weaken the attention given to the few truly significant elements \citep{shen2018reinforced,zhao2019sparse}. In the context of drug-target interactions, it shares the same problem because a drug interacts with small portions of a protein. We constrain the attention map to be sparse to encourage the model to focus on a few salient parts. We achieve this by replacing the softmax function with the sparsemax \citep{martins2016softmax} function, which considers the small weights to be unimportant and masks those values from the features when calculating the softmax-based attention.

\section{Experiments}   
In this section, we evaluate the performance of our method in a drug-target interaction task on two different datasets.
\subsection{Experimental Settings}

\subsubsection{Comparison Methods}
As previously stated, the GCA is designed to endow existing methods with interaction interpretability. Thus we compare the performance by applying our method to the following baselines:
\begin{itemize}
    \item EmbDTA: Instead of a specific feature extractor, it only uses an embedding layer. We add this to see how much the proposed method improves the performance of the task using only the interaction. It uses SMILES and FASTA to represent the drugs and targets, respectively.
    
    \item DeepDTA \citep{ozturk2018deepdta}: It uses CNNs for feature extraction layers to encode locally salient features from drugs and targets. It uses the same representations for drugs and targets as the EmbDTA.
    
    \item GraphDTA \citep{nguyen2019graphdta}: It uses GCNs to exploit the properties of the molecular graph and utilizes CNNs for proteins as in DeepDTA. For an input representation, it uses atoms in drugs and connects them through graph information (i.e., adjacency matrix). To preserve sequential information, we use a sequence-wise convolution instead of feature-wise convolution, which is similar to \cite{kim-2014-convolutional} and we empirically found that this version produces nearly identical results with that of feature-wise convolution.
    
\end{itemize}
Note that the previous methods use multi-layered feed-forward networks for an interaction layer (i.e., $\phi(\cdot)$). We replace the networks with GCA without modifying the feature extractor on previous methods. For a fair comparison, we set the nearly same number of parameters used in the interaction layers.

In our model, we use a single cross-attention layer with two heads. The training setup is the same with the previous work \cite{nguyen2019graphdta}. We have implemented the proposed method and all experiments with PyTorch \citep{paszke2017automatic} and trained the models on a single NVIDIA Tesla V100 with 32GB of RAM. 

\subsubsection{Datasets}

\begin{table}[t]
\caption{Statistics for KIBA and Davis dataset. Len. indicates the average string length of each format (SMILES, FASTA).}
\begin{adjustbox}{max width=\textwidth}
\begin{tabular}{@{}l|c|c|c|c|c|c@{}}
\toprule
Dataset & \# of Drugs & \# of  Targets & Len. Drugs (SMILES) & Len. Targets (FASTA) & \# of Interactions & Affinity range  \\ \midrule \midrule
KIBA    & 2,111       & 229            & 56.6                & 656.4                & 118,254            & {[}0.0, 17.2{]} \\ \midrule 
Davis   & 68          & 442            & 62.7                & 712.2                & 30,056             & {[}5.0, 10.8{]} \\ \bottomrule
\end{tabular}
\end{adjustbox}
\label{dataset-stat}
\end{table}

We use two benchmark datasets to evaluate the performance in the drug-target interaction task , which are \textbf{KIBA} large-scale kinase inhibitors bioactivity dataset \citep{tang2014making,he2017simboost} and \textbf{Davis} kinase binding affinity dataset \citep{davis2011comprehensive}. The statistics of these datasets are listed in Table \ref{dataset-stat}.

\subsubsection{Evaluation Metrics}
We evaluate the methods by using mean squared error (MSE) and concordance-index (C-index) score following previous works \citep{ozturk2018deepdta, nguyen2019graphdta}. MSE is the average of regression loss in the test dataset, and C-index is the probability that randomly selected pairs are correctly ordered. It is computed as follows:
\begin{equation}
    \text{C-index} = \frac{1}{N_{test}}\sum_{y_i > y_j}{\mathbbm{1}(\hat{y_i} > \hat{y_j})}
\end{equation}
where $N_{test}$ is the number of test pairs and, $y_i$, $y_j$ are affinity scores of randomly selected pairs, $\hat{y_i}$, $\hat{y_i}$ are predicted affinity scores from the trained model. $\mathbbm{1}(x)$ is a identity function that returns 1 if the statement $x$ is true, otherwise 0.

\subsection{Results}

\begin{table}[t]
\centering
\caption{Performance evaluation on KIBA and Davis. Best results are highlighted in boldface.}
\begin{adjustbox}{max width=0.75\textwidth}
\begin{tabular}{@{}l|c|c|cc@{}}
\toprule
\multirow{2}{*}{Method}       & \multicolumn{2}{c|}{KIBA} & \multicolumn{2}{c}{Davis}                      \\ \cmidrule(l){2-5} 
                               & MSE & C-index & \multicolumn{1}{c|}{MSE} & C-index \\ \midrule \midrule
EmbDTA                 & 0.342       & 0.761       & \multicolumn{1}{c|}{0.438}     & 0.811         \\
EmbDTA + GCA     & \textbf{0.311}       & \textbf{0.784}       & \multicolumn{1}{c|}{\textbf{0.393}}     & \textbf{0.832}           \\ \midrule
DeepDTA               & 0.186       & 0.865       & \multicolumn{1}{c|}{0.246}       & 0.881           \\
DeepDTA + GCA   & \textbf{0.170}       & \textbf{0.881}       & \multicolumn{1}{c|}{\textbf{0.238}}     & \textbf{0.888}          \\ \midrule
GraphDTA              & \textbf{0.151}       & \textbf{0.884}       & \multicolumn{1}{c|}{0.251}     & 0.883           \\
GraphDTA + GCA  & 0.152       & 0.883       & \multicolumn{1}{c|}{\textbf{0.242}}     & \textbf{0.891}           \\ \bottomrule
\end{tabular}
\end{adjustbox}
\label{main_exp}
\end{table}

\begin{wraptable}{r}{6cm}
\caption{Comparison with other interaction methods.}
\resizebox{0.4\columnwidth}{!}{%
\begin{tabular}{@{}l|c|c@{}}
\toprule
Method              & MSE & C-index \\ \midrule \midrule
EmbDTA       & 0.448      & 0.811           \\ \midrule
EmbDTA + DIN & 0.523      & 0.776           \\
EmbDTA + AP  & 0.439      & 0.824           \\
EmbDTA + GCA & \textbf{0.393}      & \textbf{0.832}           \\ \bottomrule
\end{tabular}%
}
\label{otherint}
\end{wraptable}

Table \ref{main_exp} shows the full comparison results in two datasets. Replacing the interaction networks with GCA consistently yields better results for most methods and datasets. Among them, EmbDTA, which has the least feature extraction parts, achieves greater performance improvement than others (roughly 9\% improvement in terms of MSE). These results suggest that constructing explicit interaction is useful to improve the performance in the task. Furthermore, we see that GCA works well with the use of other input types (i.e., characters in EmbDTA and DeepDTA, graphs in GraphDTA), and it shows that GCA can generally be applied to existing methods.

We also compare GCA with different interaction methods used in other domains. These methods are attentive pooling (AP, \cite{santos2016attentive}) and deep interaction networks (DIN, \cite{gong2017natural}). Here, we experiment on the settings of EmbDTA in Davis to solely compare the interaction layers while minimizing the influence of the feature extractor. Table \ref{otherint} presents the comparison results. Again, GCA works better than the others. In addition to achieving superior performance, we have several advantages. Compared to DIN, GCA explicitly enables the neural models to interact with a long-range relationships between drugs and target features owing to the attention mechanism. AP is the most similar method to GCA, as it also calculates attention maps and assigns weights to each feature. However, GCA adopts a multi-head and a multi-layer mechanism using the transformer, making the method more scalable by stacking multiple interaction layers and jointly attending to information from different representation subspaces at different positions \citep{vaswani2017attention}. Indeed, we observe that the second head in GCA better captures the binding sites in proteins than the first head (detailed in Section 5.2). Such differences contribute to a better performance.

\section{Analysis}

\subsection{Ablation Study}

\begin{wraptable}{r}{7.5cm}
\caption{Ablation study for attention directionality}
\resizebox{0.48\columnwidth}{!}{%
\begin{tabular}{@{}l|l|l@{}}
\toprule
Method                   & MSE & C-index \\ \midrule \midrule
GCA                      &  0.311   &   0.787      \\
GCA w/o Drug Attention   &  0.341   &   0.768     \\
GCA w/o Target Attention &  0.323   &   0.779      \\ \bottomrule
\end{tabular} %
}
\label{abl}
\end{wraptable}

In this subsection, we analyze whether cross attention is effective compared to single direction attention and which attention is more significant for performance improvement. To this end, we measure the task performance when we remove the drug ($g_{p \rightarrow d}$) or target attention ($g_{d \rightarrow p}$), respectively. Here, we use EmbDTA trained on the KIBA dataset. Table \ref{abl} lists the ablation results. As can be seen from the table, the performance decreases as we remove each directionality from GCA. Specifically, removing drug attention from targets results in the largest drop in performance. This indicates that drug attention is more significant than target attention. We also see that using both two types of attention leads to the best performance, demonstrating that cross-attention is indeed effective and that each attention complements usefully.

\subsection{Binding site prediction}

\begin{figure}[t]
  \centering
  \subcaptionbox{Exact prediction}{\includegraphics[width=.47\textwidth]{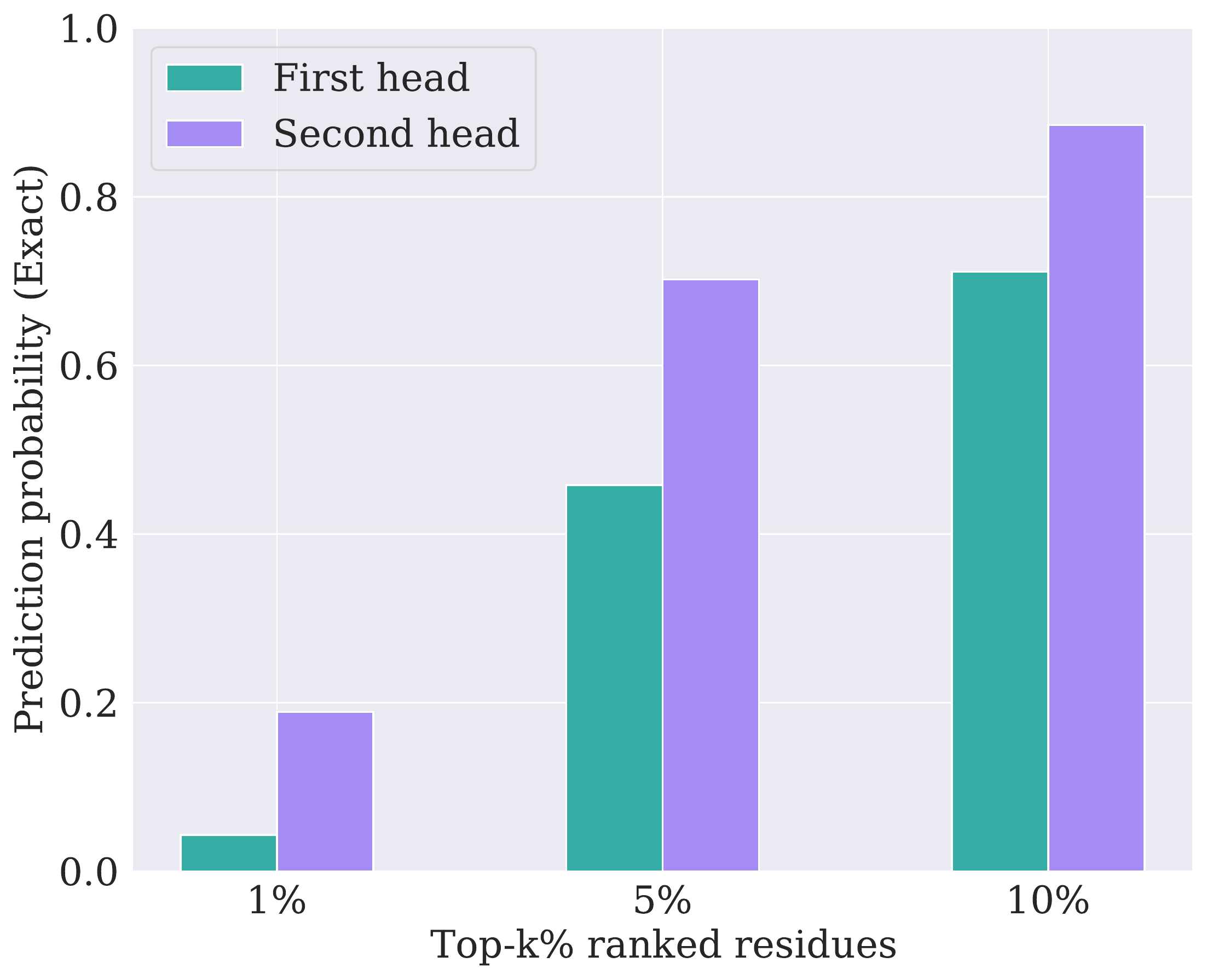}} 
  \subcaptionbox{Surrounding prediction}{\includegraphics[width=.47\textwidth]{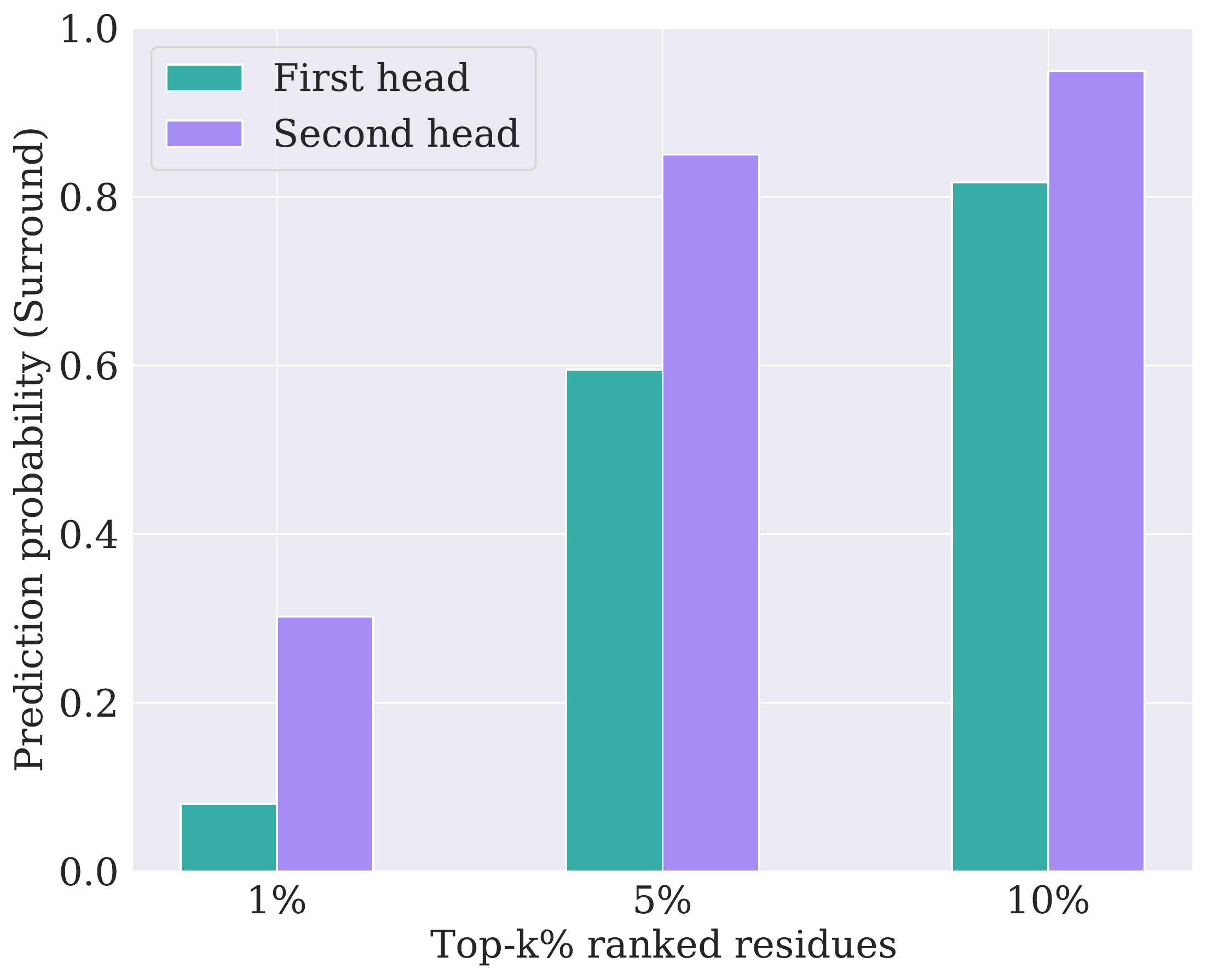}} 

\caption{Binding site prediction results. \textbf{Left} is the results of exact prediction. \textbf{Right} is the results of prediction with including nearest 1 position.}
\label{binding}
\end{figure}


The proposed method assigns attention weights to a given protein such that the highlighted residues directly affect to the task performance. Therefore, residues receiving relatively high attention can be important local residues for prediction, such as binding sites. In this subsection, we analyze how the residues that are highly ranked by the attention correspond to actual binding sites. Here, we use GraphDTA with GCA, which is trained on KIBA, and use scPDB \citep{desaphy2015sc} for evaluation, which has atomic-level descriptions of proteins, ligands, and binding sites from complex structures. It contains interactions between proteins and pharmacological ligands for 7,179 entries of Vertebrata (the number of unique proteins is 809). We measure the probability of whether the top $k$\% ranked residues include binding sites or not, and evaluate two predictions from two different heads in GCA. Figure \ref{binding} shows the prediction results. We first observe that the attention mostly finds the binding site when searching for the top 10\% residues. Furthermore, the second head from our attention better captures binding sites than first head with a large margin. This shows that our method encode different aspects of the input features through multiple heads. These results suggest that our method can provide the insight in to the interpretability of binding sites by highlighting nearby residues.

\subsection{Case Study}

\subsubsection{Interaction Interpretability }

In our method, we provide two views on the prediction results: drug attention and the target attention. In this analysis, we confirm which atoms in drugs are receiving high attention from our method. To this end, we conduct a case study of the interaction between chemicals (PubChem ID: 117793281) and protein (MAP Kinase, UniProt ID: Q9HBH9). 

\begin{wrapfigure}[]{r}{7cm}
    \centering
    \includegraphics[width=0.42\textwidth]{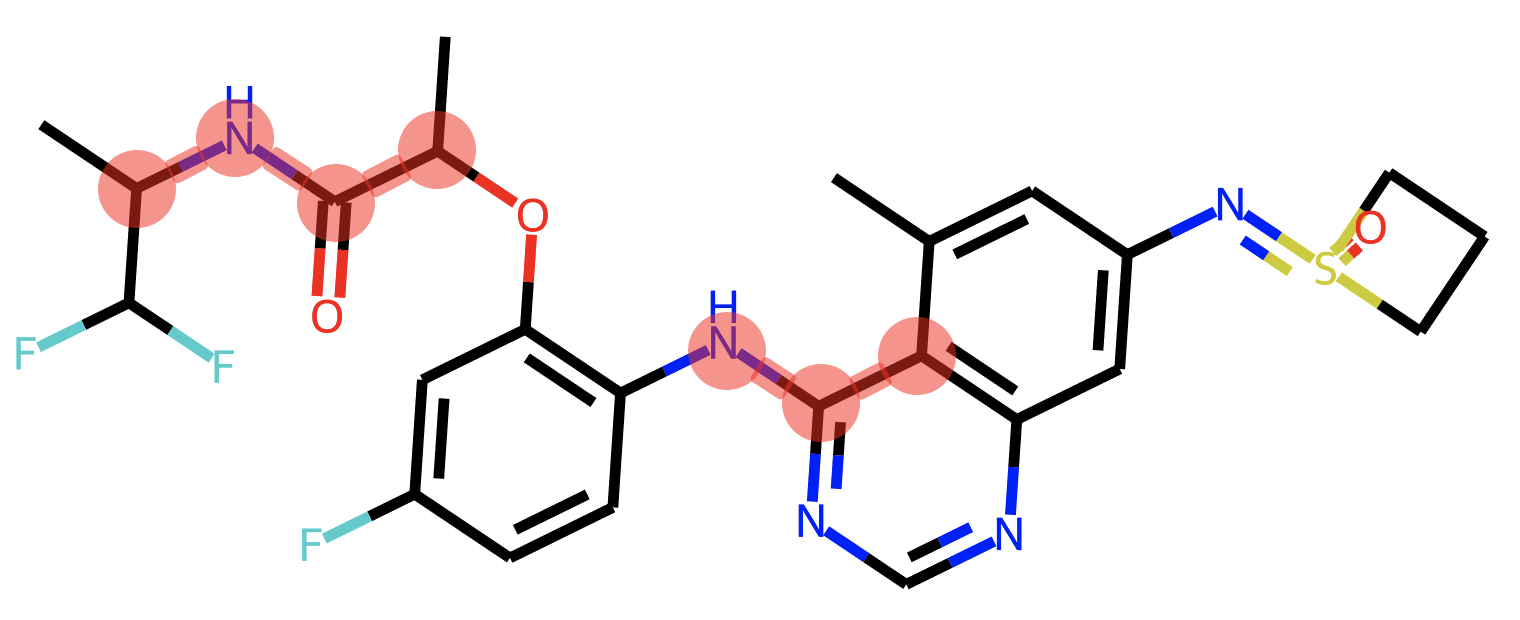}
    \caption{Highlighted atoms in the chemical (PubChem ID: 117793281)}
    \label{mols}
\end{wrapfigure}
Here, we use GraphDTA with GCA trained in KIBA and draw the top contributing atoms in Figure \ref{mols}. As can be seen from the figure, the highlighted atoms in the drug attention include several nitrogen atoms that can form hydrogen bonds with amino acids at the binding site. As we use GCNs that interact with neighbors to represent drugs, several carbons near the nitrogen have received high attention, and they can also formulate potential solvent interaction since carbon has a relatively large solvent accessible surface. Overall, the results show that the highly attended parts of the drug attention can indicate potential interaction sites in drugs. Such interpretability can help researchers find leads by displaying the regions near the interaction sites with a binding affinity score.

\subsubsection{Mutation on epidermal growth factor receptor}

\begin{table}[t]
    \centering
\caption{Rank of highlighted residues when the T790M mutation occurs}
\begin{adjustbox}{max width=0.75\textwidth}
\begin{tabular}{@{}c|c|c|ccc@{}}
\toprule
\multicolumn{3}{c|}{EGFR}            & \multicolumn{3}{c}{EGFR with T790M}                                                          \\ \midrule
Rank & Residue Index & Near Residues & \multicolumn{1}{c|}{Rank}               & \multicolumn{1}{c|}{Residue Index} & Near Residues \\ \midrule \midrule
1    & 655           & GALLL           & \multicolumn{1}{c|}{1}                  & \multicolumn{1}{c|}{655}           & GALLL           \\
...  & ...           & ...           & \multicolumn{1}{c|}{...}                & \multicolumn{1}{c|}{...}           & ...           \\
12   & 792           & \textcolor{blue}{\textbf{T}}QLMP         & \multicolumn{1}{c|}{5 (+7)}             & \multicolumn{1}{c|}{792}           & \textcolor{red}{\textbf{M}}QLMP         \\
...  & ...           & ...           & \multicolumn{1}{c|}{...}                & \multicolumn{1}{c|}{...}           & ...           \\
120  & 791           & I\textcolor{blue}{\textbf{T}}QLM         & \multicolumn{1}{c|}{\textbf{11 (+109)}} & \multicolumn{1}{c|}{791}           & I\textcolor{red}{\textbf{M}}QLM         \\ \bottomrule
\end{tabular}
\end{adjustbox}
\label{mutation}
\end{table}

Lastly, we analyze how the attention from ours is activated differently when a mutation occurs. We choose the mutation occurring at the epidermal growth factor receptor (EGFR) which is a well-known genes related to many cancer types. Specifically, we apply T790M mutation to EGFR, a well-known mutation found in most lung cancer patients. We use GraphDTA with GCA trained on KIBA. For a drug, we use Osimertinib which was developed as a third generation tyrosine kinase inhibitor targeting the T790M mutation in the exon of the EGFR gene \citep{soria2018osimertinib}. In this analysis, we compare the rank of residues in the target attention when the mutation occurs. Table \ref{mutation} lists the ranked results before and after T790M mutation, in descending order of attention weights. We denote the ranks of the near residues where mutation occurs (i.e., 791, 792 residue)\footnote{As we use CNNs to extract features from proteins, the surrounding regions of attended parts are meaningful.}. When the mutation occurs, the near residues of the mutation site are highly activated. While the increased ranking is not the most highlighted, such a drastic increase shows that GCA can react sensitively to the mutation and can be used as a guideline for identifying new drugs targeting specific mutant proteins.

\section{Conclusion}
In this paper, we have proposed a novel interpretable framework for drug-target interactions based on \textit{\textbf{gated cross attention}}, denoted as GCA. GCA is elaborately designed to make the neural model explicitly interact with drugs and targets using the attention mechanism, and it assigns weights to each feature by considering counterpart information. Because the weights directly affect the prediction, the values in the attention can serve as an interpretable factor. The experimental results support the strengths of GCA in drug-target interaction and provide pharmaceutical insights gained from GCA for binding sites and potential interactions between drugs and targets. Furthermore, with the example of T790M EGFR, we have shown that GCA reacts sensitively to the mutant protein by mainly attending the surrounding mutation site. The comprehensive results suggest that our method can provide useful cues for the predictions of deep learning-based DTI methods, and we believe that existing methods can benefit from our work.

\bibliography{mlhc}

\end{document}